# WHERE WAS COVID-19 FIRST DISCOVERED? DESIGNING A QUESTION-ANSWERING SYSTEM FOR PANDEMIC SITUATIONS

*Research Paper*

Johannes Graf, Technische Universität Dresden, Dresden, Germany, johannes.graf3@mailbox.tu-dresden.de

Gino Lancho, Technische Universität Dresden, Dresden, Germany, gino.lancho@tu-dresden.de

Patrick Zschech, Friedrich-Alexander-Universität Erlangen-Nürnberg, Nürnberg, Germany, patrick.zschech@fau.de

Kai Heinrich, Otto-von-Guericke-Universität Magdeburg, Magdeburg, Germany, kai.heinrich@ovgu.de

## Abstract

*The COVID-19 pandemic is accompanied by a massive "infodemic" that makes it hard to identify concise and credible information for COVID-19-related questions, like incubation time, infection rates, or the effectiveness of vaccines. As a novel solution, our paper is concerned with designing a question-answering system based on modern technologies from natural language processing to overcome information overload and misinformation in pandemic situations. To carry out our research, we followed a design science research approach and applied Ingwersen's cognitive model of information retrieval interaction to inform our design process from a socio-technical lens. On this basis, we derived prescriptive design knowledge in terms of design requirements and design principles, which we translated into the construction of a prototypical instantiation. Our implementation is based on the comprehensive CORD-19 dataset, and we demonstrate our artifact's usefulness by evaluating its answer quality based on a sample of COVID-19 questions labeled by biomedical experts.*

## 1 Introduction

The coronavirus disease 2019 (COVID-19) caused by SARS-CoV-2 originated from the Wuhan Seafood Market in China (Triggle et al. 2021) has been spreading worldwide and evolved to a pandemic on March 11, 2020 (Ghebreyesus 2020a). Although the mortality rate of SARS-CoV-2 is relatively lower compared to other coronaviruses, the total death toll, impact on the global economy, and each individual's lifestyle are far more severe due to its highly transmissible nature (Petersen et al. 2020) and largely asymptomatic cases (Day, 2020). One of the strategic objectives proclaimed by the World Health Organization (WHO) is to manage the "over-abundance" of information as a "massive infodemic" (Ghebreyesus 2020b) accompanies the pandemic that makes it hard to identify credible information for COVID-19-related questions, like incubation time, infection rates, symptoms, the effectiveness of vaccines, etc. (WHO 2020). While demand for health information is increasing, the readability of health information is low since scientific documents as a valuable source of verified knowledge are often written in a complex language and thus hard to understand by the general public (Szmuda et al. 2020). In a European study, 47% of respondents showed health literacy deficiency (Sørensen et al. 2015). Thus, many individuals rely on social media to fulfill their information needs as it is accessible, concise, and easy to comprehend; however, it often provides unreliable misinformation from non-scientific sources (Pennycook et al. 2020). As about 50% of the general public experiences health information overload from the COVID-19 pandemic (Mohammed et al. 2021), incumbent information systems (IS) are insufficient to address this problem. Information overload also extends to health-literate people like scientists who struggle to keep up with the volume of publications regarding COVID-19 (Brainard et al. 2020).

From a technological perspective, there are various possibilities to provide support for this dire situation. Especially the fields of *information retrieval (IR)* and *natural language processing (NLP)* have brought forth great achievements in recent years for automated text processing and the extraction of relevant content (Janiesch et al. 2021; Young et al. 2018; Zschech et al. 2019). Based on advancements in the area of deep neural networks and pre-trained universal language models, such as BERT (Devlin et al. 2019) and GPT-3 (Brown et al. 2020), there have been remarkable milestones to build efficient *question-answering systems (QAS)* without the necessity of explicitly programming such systems in time-consuming and cost-intensive development projects (Abbasiantaeb and Momtazi 2021). However, it currently lacks prescriptive design knowledge on how to design such systems from a socio-technical perspective. Thus, this paper aims to address the following research question (RQ):

**RQ:**   *How to design a question-answering system for pandemic situations to overcome information overload and misinformation?*

To answer this question, we followed a design science research (DSR) approach (Hevner et al. 2004) with the aim (i) to devise design requirements and prescriptive design principles and (ii) to develop a prototypical QAS implementation for pandemic situations. As a theoretical lens, we applied *Ingwersen's cognitive model of IR interaction* (Ingwersen 1996) to improve understanding of the design from a socio-technical perspective. The prototypical instantiation of our derived design knowledge is based on the comprehensive COVID-19 Open Research Dataset (CORD-19), a collection of more than 500,000 full-text research papers across several domains (Wang et al. 2020). For evaluation purposes, we focus in this paper on the general feasibility of the constructed QAS and demonstrate its usefulness by outlining the answer quality based on a sample of 25 COVID-19-related questions with different levels of complexity pre-labeled by biomedical experts (Möller et al., 2020).

Our paper is structured as follows. In Section 2, we introduce the preliminaries and refer to related work. Subsequently, we depict our research approach in Section 3. In Section 4, we outline our results from the design process and demonstrate the implemented prototype in Section 5. We then proceed by presenting the preliminary evaluation results in Section 6. Finally, we summarize our contribution and provide an outlook for future work in Section 7.

## 2    Preliminaries and Related Work

### 2.1    Information Overload and Misinformation in the Age of Pandemics

Compared to previous virus incidents, an exponentially growing surge of new information accompanies the current COVID-19 pandemic. The exponentially rising published research reflects this information demand: 5595 scientific papers have been released three months after SARS-CoV-2 was defined, far exceeding the numbers for previous incidents like SARS-CoV-1 (177) or MERS (28), respectively (Valika et al. 2020). Based on a US study, news consumption increased on all media platforms, particularly on social media, more than doubling from 38% to 87% of citizens (Casero-Ripolles 2020). The study emphasizes that the growth was disproportionally high among individuals with a low level of education. *Information overload* and *misinformation* that disseminates on social media, in particular, can promote anxiety (Islam et al. 2020; Liu et al. 2021) and destabilize society and institutions (Waisbord 2018). Table 1 depicts aspects of both dimensions of the infodemic.

| Dimensions | Aspects | | | |
|---|---|---|---|---|
| *Information overload* | Quantity of news and scientific papers | Quality of news and scientific papers | Time pressure for decision-makers and scientists | Limited capacity for processing new information |
| *Misinformation* | Fabricated information | | Reconfigured information | |

*Table 1.        Characteristics of the Infodemic.*

*Information overload* is a phenomenon decision-makers encounter when facing an over-abundance of information about an issue (Peabody 1965). While there is no standardized definition of information overload (Roetzel 2019), several aspects are recurrent: the *quantity* (Hiltz and Turoff 1985) implies the input of available information objects from various sources. Newer studies additionally emphasize other aspects, including *time pressure* (Pennington and Tuttle 2007; Schick et al. 1990) or the decision maker's information *processing ability* (Saunders et al. 2017), including their short-term memory, long-term memory, and cognitive capacity. Furthermore, the *quality* (Burton-Jones and Straub 2006) of information also affects decision-makers, as information may be ambiguous (Schneider 1987), redundant (Li 2017), or complex (Bawden and Robinson 2009). A study revealed that 50% of participants suffer from health information overload regarding the COVID-19 pandemic (Mohammed et al. 2021). Information overload is not limited to the general public that predominantly consumes traditional or social media (Casero-Ripolles 2020) but also affects the scientific community struggling to keep up with the influx of newly published papers (Brainard et al. 2020).

*Misinformation* refers to the spread of information that is either completely *fabricated* or *reconfigured that* existing information is twisted, recontextualized, or reworked (Brennen et al. 2020). The spread of misinformation during the COVID-19 pandemic, labeled as "infodemic" by the WHO (Ghebreyesus 2020b), is a consequence of information overload. The quantity and the complexity of the domain make it hard to distinguish credible information from misinformation, which is why the WHO declared its strategic goal to reduce the amount of false information shared (WHO 2020). Especially on social media, people find it hard to distinguish fake news from credible ones (Pennycook et al. 2020). Brennen et al. (2020) found that misinformation spreads bottom-up as well as top-down. Their study revealed that 20% of misinformation comes top-down but is responsible for 69% of engagement on social media. Examples of top-down misinformation include a well-known yoga guru and entrepreneur advertising herbal remedies as a cure against COVID-19 (Ulmer 2020) or claims regarding the ineffectiveness of face masks and social distancing (Hornik et al. 2021).

## 2.2    Question Answering Systems

Question answering (QA) is a sub-discipline at the intersection of IR and NLP that is concerned with the task of automatically answering questions posed by humans in natural language. The complex QA task requires many sub-systems and must accommodate various dimensions such as the application, the user, or the question type (Hirschman and Gaizauskas 2001). Rather than retrieving a ranked list of documents to a keyword-based query, as most IR-systems do, a QAS provides concise answers from passages in documents to queries posed in natural language (Allam and Haggag 2012). Thus, QAS are a significant advancement, especially regarding large and complex unstructured documents, while offering a viable tool to reduce information overload (Olvera-Lobo and Gutiérrez-Artacho 2010).

There are fundamentally two different classes of QAS: (i) *knowledge-based QAS* and (ii) *neural QAS* (Janiesch et al. 2021). Knowledge-based QAS require the system designer to encode formal knowledge into the QAS and are usually limited to a specific domain (Jurafsky and Martin 2009; Lan et al. 2021). They generally perform well on factoid questions but at the same time also suffer from the limitation that they are constrained to the underlying database and logical model, implying that they are inflexible and do not generalize very well. Therefore, the system designer requires in-depth knowledge about the domain to encode logic, graphs, or ontologies to form a structured QAS database (Yang et al. 2015), which may lead to a semantic gap between the system designer, content creator, and user. Furthermore, knowledge-based QAS are considered counterintuitive from a user experience perspective as they often require to work with specific query languages such as SPARQL (Abdi et al. 2018), while it is time-consuming to maintain the knowledge base with new encodings.

Neural QAS follow a different paradigm. They are not restricted to a structured and pre-defined knowledge base and therefore operate in a more flexible manner (Huang et al. 2020). From an unstructured database, these systems first retrieve a set of text passages from documents that appear relevant to the query and then extract the most likely answer(s) to return it to the user (Kratzwald et al. 2019). Neural QAS have improved significantly due to rapid advancement in deep neural network architectures, especially with the emergence of pre-trained language models such as BERT (Devlin et al. 2019) or GPT-3 (Brown et al. 2020) that can be fine-tuned for a QA task at hand. Thus, neural QAS can adapt to other databases easily as the underlying pre-trained model learns to comprehend language in general and does not rely on pre-defined knowledge bases like ontologies or graphs. Therefore, pre-trained models can be deployed for different tasks and domains, commonly known as transfer learning (Chung et al. 2018).

Most neural open-domain QAS consist at least of two main components (Zhu et al. 2021). The first one is a *retriever module* that filters the documents with IR-based methods and creates a shortlist. This step is necessary because directly analyzing all documents within large databases is computationally not efficient. Subsequently, a *reader module* as the second component is responsible for text comprehension and finding the exact answer within the list of documents received by the retriever. The reader analyses multiple text passages from the received documents and returns the top-n ranked answers. Neural open-domain QAS have achieved human-like performance, with some models even exceeding it by a slight margin having over 90% accuracy (Rajpurkar et al. 2020). Such performance is tested, for example, against SQuAD 2.0, a widely used dataset of 50,000 question-answer pairs annotated by several human workers (Rajpurkar et al. 2018). QAS in practice serve, e.g., as chatbots for customer support (Sharma and Gupta 2018), as conversational agents in medicine (Laranjo et al. 2018), or as interfaces to trigger further tasks such as placing orders (Carmel et al. 2018). Just recently, we could also observe an increasing interest in developing QAS for COVID-19 information needs (e.g., Alzubi et al., 2021; Esteva et al., 2021; Su et al., 2020). However, all these approaches have in common that they focus on architectural aspects, especially regarding technical performance, while neglecting to derive generalized prescriptive design knowledge from a socio-technical lens. We contribute to this research gap by following a socio-technically oriented DSR paradigm (Hevner et al. 2004).

## 2.3 Cognitive Model of Information Retrieval Interaction

To better understand the artifact's design from a socio-technical perspective, we applied the cognitive model of IR interactions developed by Ingwersen (1996). Since this model abstracts from purely technical details and describes the interplay between different actors and technical components during information search processes, it has been used already in previous (DSR) projects to describe the nature of IR processes (e.g., Seidel et al. 2008; Sturm & Sunyaev, 2019). For example, Sturm & Sunyaev (2019) applied the model to inform design choices for constructing *systematic literature search systems*. Since QAS show very similar properties, Ingwersen's model also provides a valuable basis for our endeavor.

Following Ingwersen's model, IR interactions generally comprise (i) *information objects* of interest (i.e., full texts, semantic entities), (ii) *technical artifacts* (i.e., IR system components and interfaces), and (iii) the *communication among human actors* (i.e., individual users and social/organization environment) on a *cognitive level*. The specific focus of this view is on users conducting IR or, as in our case, the QA task. The users' cognitive model is expressed, for example, by their individual goals, information needs, and information behaviour. By contrast, the cognitive models of information objects and technical artifacts are explications of their creators' cognitive models, i.e., system designers or authors. (Ingwersen 1996) argues that a fit between the different actors' cognitive models is essential to ensure effective IR interactions, which is also in line with closely related theories such as the *task-technology fit theory* (Goodhue and Thompson 1995) and the *cognitive fit theory* (Vessey and Galletta 1991). In other words, it is crucial to reduce or avoid inconsistencies between the cognitive models of different actors (i.e., content creators and consumers) that may lead to misunderstandings or uncertainty due to an increased interaction effort. For the given context of *providing useful answers for COVID-19-related questions*, this view has two implications. On the one hand, a system must translate target-user-specific questions into a suitable representation that reflects the consumers' original information needs. On the other hand, it needs to retrieve and provide answers from a knowledge base that is not limited to a specific format and constrains the content creators' original answer quality. We break this broad task down into more granular *design requirements* arising from the circumstances of the pandemic situation, which we further elaborate on in Section 4.

## 3 Methodology

We followed a DSR approach as a fundamental paradigm in IS research that is concerned with the construction of socio-technical artifacts to solve organizational and societal problems and derive prescriptive design knowledge (Gregor and Hevner 2013). More specifically, we followed the DSR method by Peffers et al. (2007) consisting of the six steps of (i) problem identification and motivation, (ii) definition of solution objectives, (iii) design and development, (iv) demonstration, (v) evaluation, and (vi) communication. Figure 1 depicts the adoption of the methodology to our project.

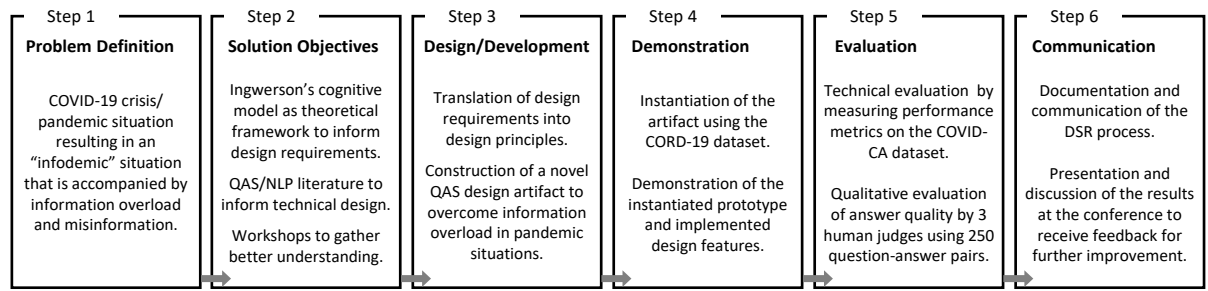

*Figure 1.        Adoption of the DSR methodology by Peffers et al. (2007).*

In the **first step**, we gathered a better understanding of the pandemic situation during the COVID-19 crisis that is characterized by many individual information needs (i.e., *COVID-19 questions*) and an abundance of generated information (i.e., *COVID-19 answers*) spread over multiple sources like social media, news, and scientific papers. As a result, we could conceptualize the "infodemic" situation, accompanied by *misinformation* and *information overload* and outlined in Sections 1 and 2.1.

In the **second step**, we specified the type of a novel artifact focusing on the class of *QAS* as a sub-type of *IR systems*. To retrieve an initial knowledge base for the definition of solution objectives, we performed several runs of literature searches. From a socio-technical perspective, we looked into previous DSR studies dealing with the design of QAS/IR systems (e.g., John et al., 2016, 2018; Robles-Flores & Roussinov, 2012; Sturm & Sunyaev, 2019; Zschech et al., 2020), which we identified via the AIS electronic library and ScienceDirect using the corresponding keywords in combination with the term "design science". From a technical perspective, we screened academic search engines like Google Scholar and Semantic Scholar for *COVID-19 question-answering systems* to inform the technical design of our artifact. However, since the first technical solutions have only recently been published or released, we broadened our scope to include current QAS surveys and conceptual papers (e.g., Abbasiantaeb & Momtazi, 2021; Breja & Jain, 2021; da Silva et al., 2020; Huang et al., 2020) to receive a general understanding about the technical realization of different QAS design alternatives.

Beyond our literature work, we also conducted workshops with a DSR scholar/NLP developer, a medical Ph.D. student, and an undergraduate economics student to retrieve a better understanding of the phenomenon. Participants were selected to capture different perspectives (i.e., information consumer vs. information provider; user vs. developer; technical vs. non-technical background; medical profession vs. non-medical novice). During the workshops, we were interested in the different facets of IR interactions related to COVID-19 question-answering. On this basis, we could gather a better under-standing of diverse search behaviours (e.g., ad-hoc vs. task-driven), individual information needs (yes/no vs. factoid questions), and the usage of commonly used IR systems (e.g., Google search, social media, PubMed). We incorporated these additional sources of information into our design process.

In the **third step**, we designed our novel QAS artifact. For this purpose, we derived *design principles* to adequately meet our specified design requirements (cf. Section 4). We translated the principles into concrete *design features* to implement a prototypical instantiation for further demonstration purposes as part of the **fourth step** (cf. Section 5).

In the **fifth step**, we evaluated our artifact's general feasibility by assessing its answer quality quantitatively while also reflecting the results on a qualitative basis. Finally, we finish the first design cycle by communicating our results in this current form as part of the **sixth step**. Since the DSR methodology is an iterative process (Peffers et al. 2007), we will use this conference format to get feedback for subsequent iteration cycles planning to introduce our QAS to a broader audience and conduct further studies to assess our findings in socio-technical user-centric experiments.

# 4    Design and Development

In this section, we present our derived design knowledge in the form of *design requirements*, *design principles,* and *implemented design features* that emerged from our design and development process. Figure 2 provides an overview of the relations between the individual elements, which we further elaborate on in the corresponding subsections. We demonstrate the design features in Section 5.

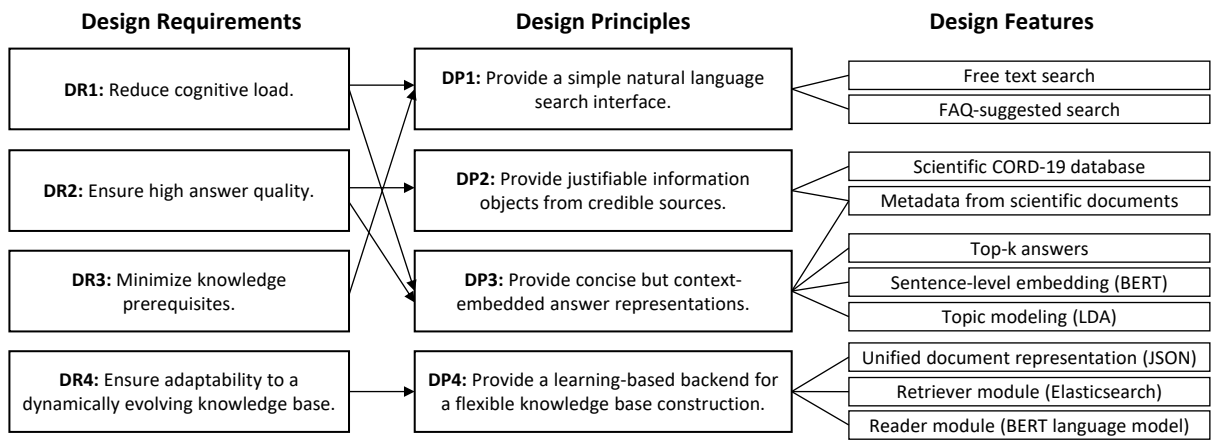

*Figure 2.        Relations between design requirements, design principles, and design features.*

## 4.1    Design Requirements

Informed by Ingwersen's cognitive model of IR interaction and the characteristics of the pandemic/infodemic situation, we identified four design requirements (DRs) for grounding our design process.

The first DR addresses the *cognitive load* of information consumers who search for pandemic-related information. Scientists and the general public face an information overload due to large volumes of data from various sources. For researchers, keeping up with the current level of knowledge, drawing relevant insights, and making decisions have become increasingly difficult due to the frequency of newly released information (Brainard et al. 2020). Thousands of COVID-19 related research papers are published and added to domain-specific databases like CORD-19 weekly, including about 500,000 scholarly articles in total (Wang et al. 2020). By contrast, the general public's information overload stems from the extensive media coverage by traditional and social media (Casero-Ripolles 2020). Due to the prevalence of health information overload regarding the COVID-19 infodemic (Mohammed et al. 2021), an appropriate artifact thus needs to reduce the cognitive effort that individuals experience to fulfill their information needs while making scientific information manageable to the general public by decreasing the complexity of information objects. To summarize, we formulate the following DR:

**DR1:** *Reduce the cognitive load of the user when searching for information during pandemics.*

The second DR reflects on data veracity and can be directly related to one of the strategic objectives declared by the WHO, i.e., to reduce the spread of *misinformation* (Ghebreyesus 2020b). In this context, Pennycook et al. (2020) found in a study with 1,700 U.S. adults that people share false information about COVID-19 as they do not sufficiently think about whether the content and the source are accurate. The WHO deems information provided from reliable sources as valid tools to curb misinformation that is spreading predominantly on social media (WHO 2020). Against this background, we formulate the following second DR for our artifact:

**DR2:** *Ensure high answer quality from credible sources to reduce the infodemic's negative consequences in terms of misinformation.*

The third DR addresses the *necessary knowledge prerequisites* that constitute a core aspect within the cognitive model of IR interactions since they can lead to a large gap when searching for information about an unfamiliar topic such as COVID-19. Information from social media is easily comprehensible but not credible and more likely subject to misinformation (Pennycook et al. 2020). On the other hand, information from scientific sources is trustworthy but not very accessible to the general public as they often lack the required e-health literacy (Norman and Skinner 2006). About 47% of EU citizens have insufficient health literacy (Sørensen et al. 2015), which renders e-health resources inaccessible to large segments of the population and facilitates the spread of misinformation (Cuan-Baltazar et al. 2020). Likewise, even highly skilled experts from the medical domain may struggle with existing tools that require a higher degree of computer literacy, e.g., understanding IR techniques that work with complex operators for search queries (Vanopstal et al. 2013). Thus, a system should reduce the complexity for obtaining information objects from the medical domain to make scientific content more accessible to a broad audience, which is why we formulate the following DR:

**DR3:** *Minimize knowledge prerequisites to make pandemic-related information more accessible.*

The fourth DR is related to the *dynamics* of information needs and produced content. Due to the strong impact of the crisis and resulting demand, scholars from multiple disciplines contribute to a global knowledge base. Particularly in the medical and pandemic domain, it is necessary to consider numerous sub-fields, such as virology or public health relations, to provide adequate answers. In this context, Montani et al. (2021) exemplify the importance of multi-disciplinary approaches to treat long-term implications of SARS-CoV-2 patients. Considering the multi-disciplinarity of the phenomenon, it is worth noting that scientific consensus and uniformity regarding a specific topic are generally scarce (Lopatovska and Arapakis 2011). Likewise, the current state of the art in many disciplines changes rapidly, which is reflected, for example, by new vaccines or medical treatments that are being tested and approved in record time. To take account for these crucial dynamics, including consensus and dissent scientific developments from multiple sub-disciplines, we formulate the following DR:

**DR4:** *Ensure adaptability to a dynamically evolving knowledge base to reflect the current state-of-the-art from a multi-disciplinary field.*

## 4.2    Design Principles

We address the defined requirements and derive a set of design principles (DPs) providing solution-related statements on *how* (materiality-oriented view) and *for what* (action-oriented view) to build our intended artifact (Chandra et al. 2015). To this end, we formulated an initial set of DPs and iteratively refined them by reflecting on our design activities in conjunction with our knowledge acquisition process (i.e., literature search and workshops). In the following, we outline our final set of principles.

**DP1:** *Design the system with a simple natural language search interface so that users with a wide variety of knowledge backgrounds can use the system with minimal effort.*

Our first DP addresses two DRs simultaneously: to reduce the user's cognitive load (DR1) and to minimize knowledge prerequisites (DR3). Compared to IR systems in other domains, a QAS providing pandemic-related information covers many user types, which differ significantly in background knowledge, information needs, and information-seeking behaviour. On the one hand, a spike in topic-related publications signals a strong interest from (i) *professional users* like academia or medical staff. This user group's characteristic information needs are typically well-defined, mitigating uncertainty and allowing conscious navigation (Ingwersen 1996). Hence, the QAS design must consider comprehensive prerequisites and provide an adequate and persuasive interface (Mayer and Moreno 2003; Tawfik et al. 2014). On the other hand, the COVID-19 pandemic also polarized the mainstream media, attracting interest from (ii) *non-professional users* (Casero-Ripolles 2020). Their information needs are comparatively ill-defined, leading to unconscious search behaviour like browsing (Ingwersen 1996), which requires a simple entry point. Although conceptualizing two discrete user groups allows for a convenient characterization, there is a continuous spectrum between the two extreme cases in real life. The user variety implies a gap of prerequisites, vocabulary, and syntactical structures between different information consumers and information providers. Therefore, a system must map questions and answers to a shared semantic space to overcome this disparity without special syntactical requirements for the user, such as a specific query language (e.g., for PubMed queries) or medical terminus. Designing a QAS with the functionality to process natural language enables this mapping and mitigates the cognitive "free fall" (Ingwersen 1996) to the symbolic level while retaining fractions of meaning from various input types.

**DP2:** *Provide the system with the ability to output justifiable information objects from credible scientific sources so that users can evaluate an answer's quality concerning the given question.*

The second DP targets the demand for high-quality answers for pandemic-related information needs (DR2). In this regard, the WHO emphasizes the importance of trustworthy information (WHO 2020). Using scientific databases like CORD-19 (Wang et al. 2020) is particularly beneficial in this context to establish a credible ground truth since their reliability is easily assessable if the responsible curator institutions are well-known, e.g., the National Library of Medicine. From such collections, answers can be extracted that reflect the information providers' (i.e., researchers, authors) original message and intention without distorting the content through additional information brokers. The format of the corresponding information objects, i.e., scientific papers, allows to consider and forward the context of a specific answer. Providing context is decisive since an answer's quality is dependent on subjective justification (Hirschman and Gaizauskas 2001), which means that the system should provide such justifiable information to the answer, allowing for a proper evaluation based on its origin.

**DP3:** *Provide the system with the functionality to extract concise but context-embedded answer representations so that users can estimate the answer quality with minimal effort.*

The third DP aims at reducing a user's cognitive load (DR1) while also ensuring a high answer quality (DR2). The previous DP2 already considers that relevant information stems from a trustworthy knowledge base, fulfilling DR2. However, knowing where answers come from is not sufficient in terms of quality as the user additionally needs to place these answers into context. Thus, the system must provide functionalities to enrich an answer with further contexts, such as additional information about the paragraphs and documents from which the answers are extracted. Likewise, our workshops revealed that considering multiple answer alternatives from different documents provides further insights, allowing the user to better assess answer quality through data source triangulation (Denzin 2017; Patton 1999) by comparing answers from different angles. In this regard, the system could improve the user's confidence in the answer when it is confirmed through multiple sources, increasing information validity (Carter et al. 2014). Information validity is particularly relevant for pandemic-related questions to reflect on scientific consensus and dissensus in a dynamically evolving environment. However, one needs to refrain from providing too much information with context. On the one hand, it is a question of obtaining a precise answer to limit the information overload and thus the cognitive load of the user (DR1), but on the other hand, it is also a question of obtaining sufficient information to guarantee a high answer quality (DR2). Thus, DR1 and DR2 are in a somewhat conflicting relation. Therefore, DP3 demands careful consideration, specifically when deciding which and how to implement corresponding design features (cf. Section 5).

**DP4:** *Provide the system with a learning-based backend for a flexible knowledge base construction so that users have access to current knowledge from a multi-disciplinary field.*

The fourth DP addresses the need to adapt to a dynamically changing knowledge base (DR4). The scientific community's knowledge base on COVID-19 is steadily evolving at a rapid pace as researchers from many disciplines contribute to its further development. Published in various databases, outlets, and formats, the system must cope with this heterogeneity to create a unified and standardized data source collection. The artifact needs to transform data to knowledge via information (Ackoff 1989) and thus, construct a knowledge representation that reflects the current state-of-the-art, considering perspectives from a multi-disciplinary field. In this regard, the system must be flexible to continuously adapt to changes in the information environment to retain a representative knowledge base, allow scalability, and ensure domain variability. The frequency of newly published information regarding pandemics renders systems that manually need to encode logic or ontologies to construct the knowledge base infeasible. Choi et al. (2016) stress this in their work to low-dimensional representations of medical concepts and emphasize the scalability of a common latent space since it significantly mitigates the strenuous task of manual feature engineering and mappings for different contexts. Therefore, DP4 emphasizes the need to incorporate learning-based retrieval and QA technologies (cf. Section 2.2) to construct an up-to-date knowledge base dynamically. This approach also corresponds to the benefits of automated analytical model building by employing data-driven machine learning algorithms as opposed to manually defining and programming relationships and decision logic into knowledge-based (retrieval) systems through handcrafted rules (Goodfellow et al. 2016; Heinrich et al. 2021; Janiesch et al. 2021; Mitchell 1997).

## 5 Prototype and Demonstration

In this section, we briefly describe our instantiated prototype and its implemented design features, which we derived from the design principles. As depicted in Figure 3, the system's architecture is composed of a frontend and a backend.

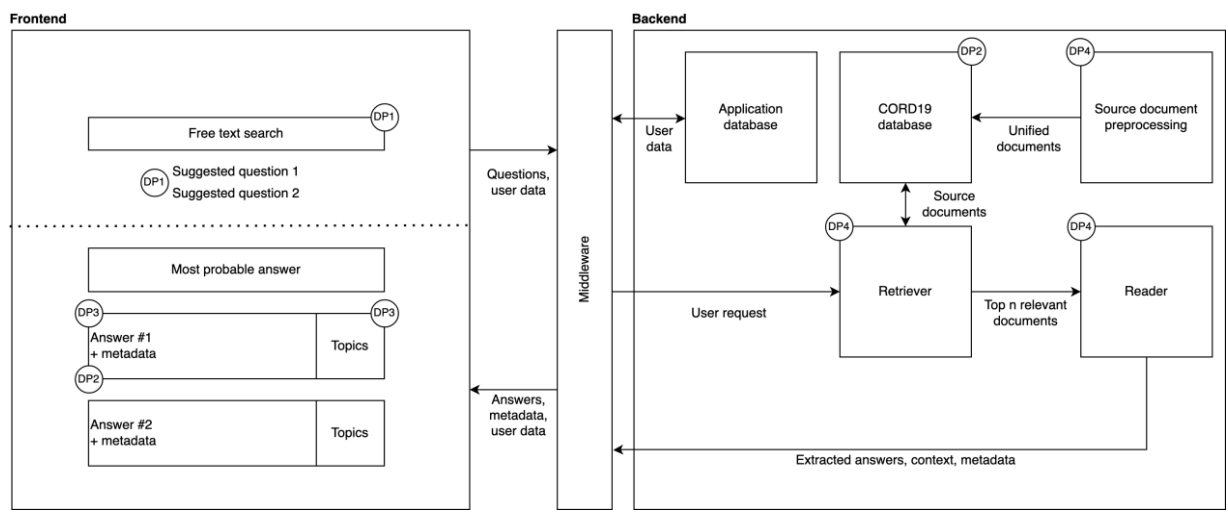

*Figure 3.        The instantiated architecture of the COVID-19 question-answering system.*

The frontend directly supports a user's IR interaction through a graphical user interface, while the backend implements the application and processing logic. A REST interface middleware handles the communication between both components. Our implementation is available to download online at a Github repository[1].

Through the frontend, users can interact with the QAS via two implemented features: firstly, by selecting from *ready-made frequently asked questions* (FAQ) suggested by the WHO, or secondly, by formulating *free text questions*. This functionality generally serves **DP1** as it minimizes knowledge prerequisites by proposing relevant questions and accepting requests in natural language to reduce semiotic gaps and the cognitive effort when formulating questions. Likewise, it supports different user groups within the spectrum of ill-defined information needs (FAQs for explorative browsing) vs. well-defined ones (free text search for intentional questions) shown in Figure 4.

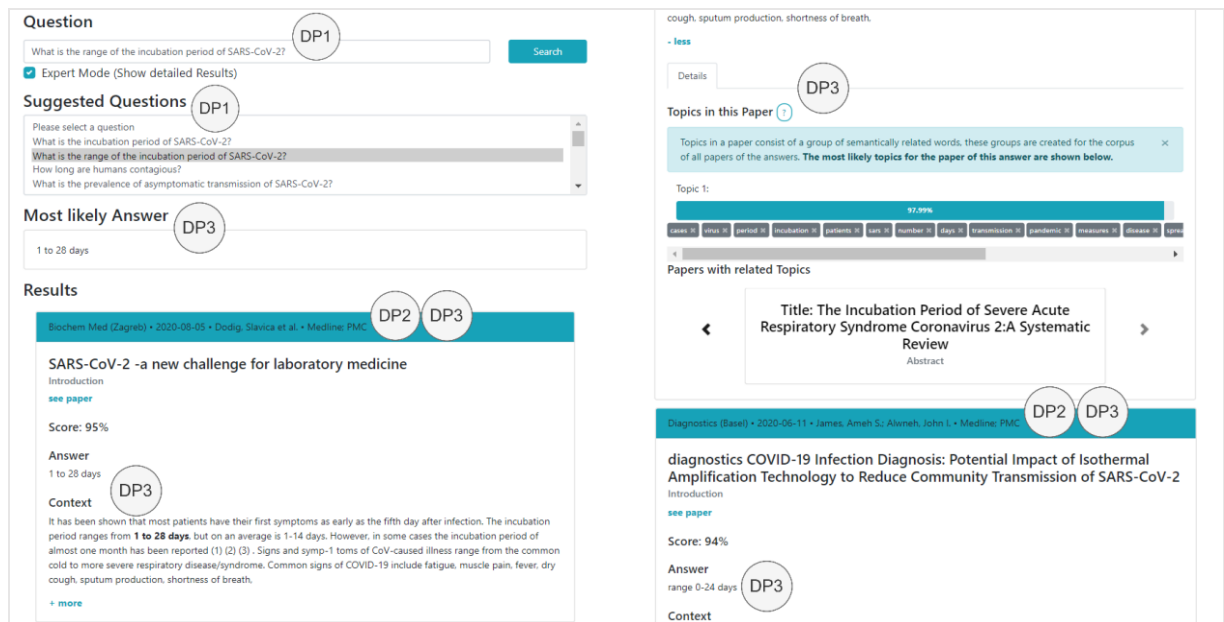

*Figure 4.        Search and top1 answer with context (left), topic model and top2 answer (right).*

After receiving user input, the backend processes the request. The backend architecturally consists of a *retriever* and a *reader* module as part of a learning-based neural QAS (cf. Section 2.2) that guarantees the flexible knowledge base construction in dynamically changing environments, as defined by **DP4**. A Haystack pipeline (Deepset 2021a, 2021b) serves as the overall QAS framework. We implement retriever through a sparse Elasticsearch retriever (Deepset 2021c) that provides the document relevance functionality. The reader employs BERT (Devlin et al. 2019); a language model that was pre-trained with vast amounts of diverse data, such as the Wikipedia corpus, which allows it to grasp the general semantics and terminology of the English language and thus ensures scalability to various contexts, to either historical data or different subdomains. The particular BERT language model deployed in our instantiation was pre-trained and distilled by Turc et al. (2019) and fine-tuned by Mezzetti (2021a, 2021b) to adapt to the medical domain and increase performance on the QA task. To ensure our QAS's multi-sourcing capability, we also implemented a pre-processing module for a *unified document representation* via JSON to treat documents of different sources equally.

---

[1] https://github.com/Covid19-QAS-DSR/ECIS2022

The database that feeds our QAS' knowledge base ingests a sample of 300,000 documents as part of the *scientific CORD-19 dataset* that contains a universal collection of over 500,000 papers regarding COVID-19 and coronaviruses across several domains (Wang et al. 2020). Well-known organizations such as the Allen Institute of AI, the National Library of Medicine, and several corporate partners curate and maintain CORD-19 to ensure credible ground truth data from multiple sources (e.g., variety of publishing institutions), directly contributing to **DP2**. In this vein, we implemented a design feature that provides all relevant *meta-data* (e.g., outlet, author, institution, publication date) from the retrieved documents, allowing users to justify the answer's origin and thus its quality and relevance.

After processing a query in the backend, the user receives the response (cf. Figure 4). Addressing **DP3**, the system displays answers in multiple levels of granularity to ensure a concise but context-enriched representation. The core output is the presentation of the most likely answer only. Further, it is possible to show the *top-k answers* (ranked by their probability scores) from additional documents/sources to verify their quality. Besides a precise answer (highlighted in bold letters (cf. left screen Figure 4), our QAS also outputs the surrounding context embedding the answer. The applied language model realizes these *sentence-level embeddings* as an architectural component. More context is also provided by the additional *meta-data* as already mentioned above. Beyond that, we implemented a *topic model* using latent Dirichlet allocation (Blei et al. 2003) that we trained with answers from our system. The resulting topic model enables the user to further grasp the context of the answer by summarizing the central theme of a corresponding document with meaningful keywords. This function also aims at reducing the cognitive effort of the user when assessing the relevance of a scientific paper since it drastically reduces the conveyed information to a relevant minimum.

# 6  Evaluation

## 6.1  Technical Evaluation and Model Selection

In the first part of our artifact's evaluation, we provide the results from a technical assessment as a crucial step for computational DSR projects (Rai 2017). For this purpose, we used COVID-CA, a QA dataset with 2,019 question/answer pairs annotated by 15 biomedical experts (Möller et al. 2020). The annotations stem from 147 research articles of the CORD-19 dataset and use a format similar to the widely known SQuAD2.0 dataset (Rajpurkar et al. 2018). However, COVID-QA differs in having, on average, a higher document length (6,118.5 vs. 152.2 tokens) and longer answers (13.9 vs. 3.2 words). Table 2 depicts two examples of the question/answer pairs and their context from COVID-QA.

Given the modular architecture of our artifact, we conducted two stages of computational experiments: One for the evaluation and selection of the retriever and one for the reader. For both phases, we tested several alternative approaches and measured their performance. Table 3 provides an overview of some of the evaluated models. For the *retriever*, we compared two dense retrievers (RT1, RT2) and one sparse retriever (RT3), using recall, mean average precision (MAP), mean reciprocal rank (MRR), and average retrieve time per document as well-known IR metrics (Manning et al. 2008; Voorhees 2001) to evaluate their performance. The sparse Elasticsearch retriever (RT3) achieved solid performance results and significantly outperformed both dense retrievers in all metrics, which is why we integrated this model into our artifact and continued with the assessment of a suitable reader.

| Question | Context | Answer |
|---|---|---|
| Where did SARS-CoV-2 originate? | "On December 29, 2019, clinicians in a hospital in Wuhan City, China, noticed a clustering of cases of unusual pneumonia (...) Within four weeks, by January 26, 2020, the causative organism had been identified as a novel coronavirus (…)" | Wuhan City, China |
| What molecules have been shown to hinder T cell responses to viral infections? | "(…) Excessive reactive oxygen species (ROS) was shown to hinder T cell responses to viral infection [36] and ROS accumulation was detected in autophagy-deficient effector T cells (…)" | Excessive reactive oxygen species (ROS) |

*Table 2.        Examples of the test data from COVID-QA.*

For the reader, we used a distilled version of BERT (Turc et al. 2019) with two different fine-tuning strategies and compared the results with the following metrics: accuracy, exact match (EM), F1-score, and total time required to extract answers out of retrieved documents for all questions. In both strategies, the pre-trained language model was fine-tuned with unsupervised training on the CORD-19 data and supervised training on the SQuAD2.0 collection (RD1, RD2). Additionally, RD1 leverages the labeled CORD-19QA dataset (Mezzetti 2020). However, as seen in Table 3, further training decreased the performance for all metrics, excluding the required reading time, as the language model lost some generalizability through the additional supervised training on the domain-specific QA dataset. For this reason, we excluded this approach and incorporated RD2 into our artifact.

| Retriever – Model Type | Recall | MAP | MRR | Time$_{average}$ |
|---|---|---|---|---|
| RT1: Dense passage retriever (Deepset 2021b; Karpukhin et al. 2020) | 35.48% | 20.23% | 20.31% | .0334s |
| RT2: Embedding retriever (Deepset 2021d; Reimers and Gurevych 2019) | 17.83% | 8.82% | 8.82% | .0526s |
| RT3: Elasticsearch sparse retriever (Deepset 2021c) | **80.49%** | **60.83%** | **60.87%** | **.0095s** |
| Reader – Model Type | Accuracy | EM$_{k=1}$ | F1 | Time$_{total}$ |
| RD1: BERTsmall_CORD19_CORD-19QA (Mezzetti 2021c) | 74.34% | 18.45% | 51.18% | 479.42s |
| RD2: BERTsmall_CORD19_SQuAD2.0 (Mezzetti 2021b) | **76.77%** | **21.51%** | **53.02%** | **498.92s** |

*Table 3.        Results of the technical evaluation for the retriever (top) and the reader (bottom).*

## 6.2     Evaluation of Answer Quality

In the second part of our evaluation, we tested our final artifact's answers quality on a more qualitative level. For this purpose, we recruited two human judges who independently scored the correctness of top$_{k}$=10 answers of our QAS for 25 questions of varying complexity (cf. examples from Table 2 vs. Table 4), resulting in a total of 250 question/answer pairs. The questions with corresponding ground truth (GT) answers are taken from the labeled COVID-QA dataset (Möller et al. 2020).

| Answer category | Question | Ground truth answer | System answer |
|---|---|---|---|
| Exact match (i) | What kind of test can diagnose COVID-19? | rRT-PCR test | PCR-based tests |
| Partial match (ii) | Where did SARS-CoV-2 originate? | Wuhan City, China | China |
| Non-GT match (iii) | Where did SARS-CoV-2 originate? | Wuhan City, China | primary host bats |
| False answer (iv) | What kind of masks are recommended to protect healthcare workers from COVID-19 exposure? | N95 mask | Surgical masks |

*Table 4.        Answer categories with examples to assess our QAS's answer correctness.*

We defined an answer's correctness on four different levels shown in Table 4 with examples. An exact match (i) is semantically identical to the GT and encompasses all answer dimensions. Therefore, exact matches (i) do not necessarily require 100% syntactical overlap. A partial match (ii) includes a subset of the ground truth (e.g., either "Wuhan" or "China" but not both). A non-GT match (iii) is a factually correct statement but differs from the GT within the COVID-QA answers. Finally, an answer is false (iv) when none of the previous classes apply (e.g., surgical masks are not fit for healthcare workers since they do not provide sufficient self-protection).

Based on the defined categories, the two judges classified all 250 answers. After an independent run, they reached an inter-rater agreement of 69.6%. Therefore, we conducted a second run in which they discussed their difference of opinions to resolve them. After this step, they reached an agreement of 87.2% with 32 open disagreements due to a lack of medical knowledge. For this purpose, we consulted a third judge with a medical background to resolve the remaining issues. See our repository (cf. Section 5) for a detailed overview of the results with all 250 question-answers pairs, GT data, and the judgments over all three iterations. Using the final assessment, we could calculate cumulated percentages of correct answers for each category (i-iii) given the top$_{k}$ answers as shown in Table 5. For example, considering the top$_{k}$=3 answers, 52% have at least one exact match (i), 80% have one (or more) partial match, and 92% have at least one non-GT match (iii). The curve already shows saturation at top$_k$=5 with a satisfying result of 72% exact matches and 100% factually correct statements, which is a solid performance given the underlying complexity of the question-answering task.

| Metric | k =1 | k=2 | k=3 | k=4 | k=5 | k=6 | k=7 | k=8 | k=9 | k=10 |
|---|---|---|---|---|---|---|---|---|---|---|
| Exact match (i) | 16% | 36% | 52% | 64% | 72% | 72% | 72% | 72% | 72% | 72% |
| Partial match (ii) | 44% | 64% | 80% | 80% | 88% | 88% | 88% | 92% | 92% | 92% |
| Non-GT match (iii) | 68% | 84% | 92% | 92% | 100% | 100% | 100% | 100% | 100% | 100% |

*Table 5.        Evaluation results of the qualitative answer assessment.*

# 7        Discussion and Concluding Remarks

Ågerfalk et al. (2020) emphasize the central role of IS to alleviate pandemic situations both long-term and short-term and called for more research on examining and designing useful IS artifacts. We followed this call and conducted a DSR project designing a question-answering system for COVID-19 information needs to fight the negative consequences of the infodemic. By drawing on the cognitive model of IR interaction and modern technologies from the field of NLP and neural QAS, we derived prescriptive design knowledge that can serve as a blueprint to inform the design of IR systems in the context of pandemics, not only for the current crisis but also for potential events in the future. Furthermore, we provided an instantiated prototype that yields promising evaluation results and can support medical and non-medical users alike to find answers for their COVID-19-related questions once we make the system publicly available to a broader audience.

However, our research is not free of limitations. Due to space restrictions, we could not introduce and demonstrate all implemented design features. These include, for example, workflow integration functionalities to support medical users in task-related IR interactions or bookmark options for storing the queried results, which we will present in later work. Another limitation concerns the evaluation of the developed QAS. At this stage, we focused on our artifact's overall feasibility and answer quality closely related to requirement DR2. Thus, in the next DSR cycle, we plan to conduct user-centric studies and evaluate to what extent our QAS can also minimize cognitive load (DR1) and allow IR interactions with minimal knowledge prerequisites (DR3). For this purpose, we are currently preparing laboratory experiments with different user groups to compare our artifact's usefulness against commonly applied IR tools like PubMed and Google Search/Scholar.

Likewise, in a more detailed study, we consider demonstrating how well our QAS can ensure adaptability in a dynamic knowledge environment (DR4). For this purpose, we conducted experiments with different backup versions of the retrieved knowledgebase before and after COVID-19 vaccines were introduced to assess how well the quality of responses improves over time. Specifically, we asked our QAS *"are there vaccines against COVID-19?"*. A system with a dataset from October 2021 refers to currently available mRNA and vector vaccines, while the same QAS with a dataset from October 2020 answers that there are no specific vaccines and antivirals yet, but there are clinical trials for these drugs. These initial results show how the system reflects the respective state-of-the-art of the underlying database. In future research, we will discuss this aspect in more detail.

Furthermore, we plan to add supplementary data sources to our artifact in subsequent design cycles. Our QAS currently incorporates an extensive collection of research papers from the CORD-10 database, covering various domains. However, we believe that not all information needs, especially critical questions from COVID-19 skeptics, might be successfully and satisfactorily answered by our QAS based on peer-reviewed research papers. Thus, we will look at additional trusted data sources, e.g., those from the WHO, to extend the functionality of our system. Nevertheless, we are confident that further data sources can be easily integrated with minimal effort, as the innovative learning-based architecture of our neural QAS allows for easy expansion of the knowledge base through transfer learning capabilities.